\documentclass[conference]{IEEEtran}
\IEEEoverridecommandlockouts
\usepackage{algorithmic}
\usepackage{booktabs}
\usepackage{subcaption}
\usepackage{multirow}
\usepackage{booktabs}

\usepackage[font=small]{caption}
\usepackage{enumitem}
\setlist{nosep}

\usepackage{cite}
\usepackage{amsmath,amssymb,amsfonts}
\usepackage{graphicx}
\usepackage{textcomp}
\usepackage{xcolor}
\usepackage[ruled,vlined]{algorithm2e}
\def\BibTeX{{\rm B\kern-.05em{\sc i\kern-.025em b}\kern-.08em
    T\kern-.1667em\lower.7ex\hbox{E}\kern-.125emX}}

\begin{document}

\bstctlcite{BSTcontrol}

\title{SDFP: Speculative Decoding with FIT-Pruned Models for Training-Free and Plug-and-Play LLM Acceleration}

\author{
Hanyu Wei$^{*}$$^{1}$\thanks{$^{*}$Work done during an internship at AMD.},
Zunhai Su$^{1}$, Peng Lu$^{2}$, Chao Li$^{2}$, Spandan Tiwari$^{2}$, Ashish Sirasao$^{2}$, Yuhan Dong$^{\dagger1}$\thanks{$^{\dagger}$Corresponding author.}\\
$^{1}$Tsinghua University,\quad $^{2}$Advanced Micro Devices, Inc.\\
\{wei-hy23, zh-su23\}@mails.tsinghua.edu.cn,\ dongyuhan@sz.tsinghua.edu.cn\\
\{Peng.Lu2, chao.li, Spandan.Tiwari, Ashish.Sirasao\}@amd.com
}

\maketitle

\begin{abstract}
Large language models (LLMs) underpin interactive multimedia applications such as captioning, retrieval, recommendation, and creative content generation, yet their autoregressive decoding incurs substantial latency. Speculative decoding reduces latency using a lightweight draft model, but deployment is often limited by the cost and complexity of acquiring, tuning, and maintaining an effective draft model. Recent approaches usually require auxiliary training or specialization, even training-free methods incur costly search or optimization. We propose SDFP, a fully training-free and plug-and-play framework that builds the draft model via Fisher Information Trace (FIT)-based layer pruning of a given LLM. Using layer sensitivity as a proxy for output perturbation, SDFP removes low-impact layers to obtain a compact draft while preserving compatibility with the original model for standard speculative verification. SDFP needs no additional training, hyperparameter tuning, or separately maintained drafts, enabling rapid, deployment-friendly draft construction. Across benchmarks, SDFP delivers 1.32×–1.5× decoding speedup without altering the target model's output distribution, supporting low-latency multimedia applications.
\end{abstract}


\begin{figure*}[h]  
    \vspace{-20pt}
    \centering
    \includegraphics[width=\textwidth]{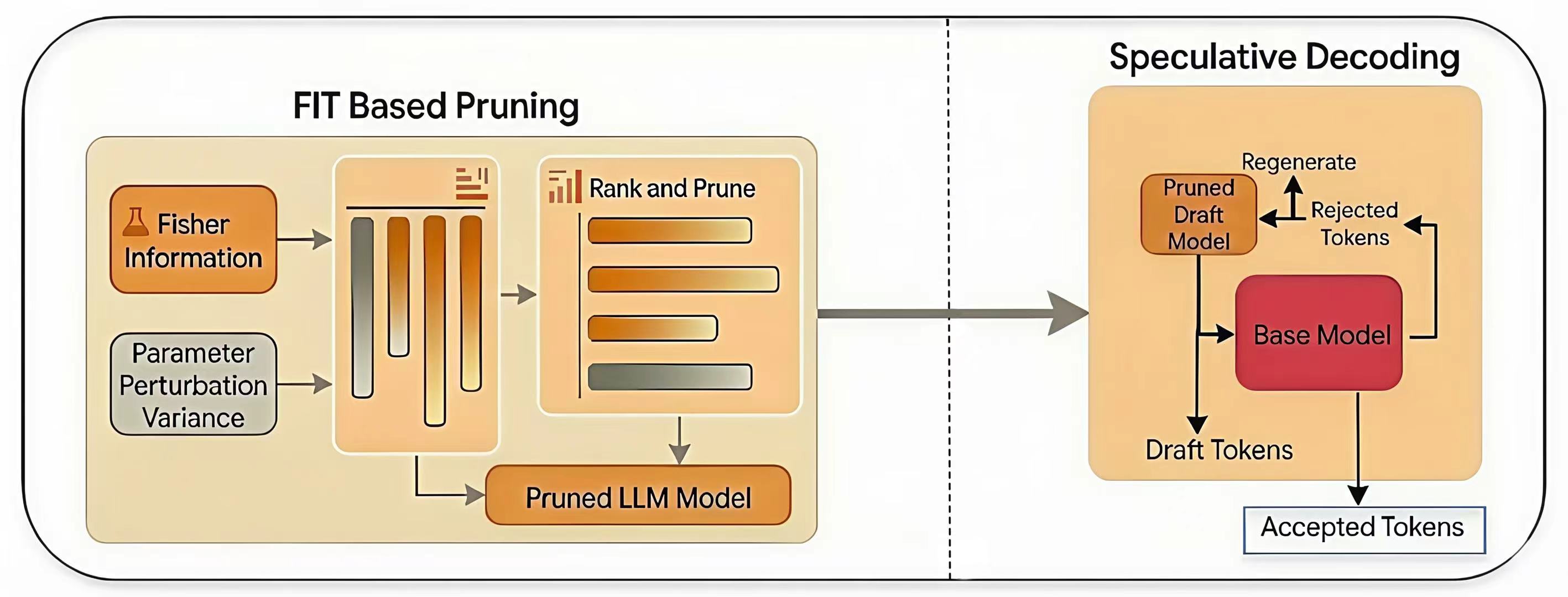}
    \caption{
        \textbf{Overview of the SDFP framework.}
        SDFP integrates \textit{FIT-based pruning} and \textit{speculative decoding} to enable training-free and plug-and-play acceleration of LLMs.
        In Stage 1, FIT scores, computed using Fisher Information and Parameter Perturbation Variance, are used to rank layer sensitivity and prune redundant layers, resulting in a compact draft model.
        In Stage 2, the pruned model acts as the draft to generate speculative tokens, which are then verified by the base model.
        Accepted tokens are committed to the output, while rejected ones are regenerated, achieving efficient decoding without retraining.
        }
    \label{fig:sdfp_overview}
    \vspace{-15pt}
\end{figure*}
\section{Introduction}
Large Language Models (LLMs) have demonstrated remarkable capabilities across a wide range of downstream tasks and are increasingly serving as key components in modern multimedia systems, such as multimodal content understanding, interactive media generation, and real-time human–computer interaction~\cite{openai2023gpt4,dubey2024llama3}. However, the autoregressive decoding paradigm, where tokens are generated sequentially, introduces substantial computational and latency overheads, which become critical bottlenecks in latency-sensitive and high-throughput multimedia applications as model scales continue to grow. To mitigate this limitation, speculative decoding (SD) has emerged as a promising technique for lossless inference acceleration~\cite{xia2023speculative,leviathan2023fast}, where a lightweight draft model predicts multiple future tokens that are then verified in parallel by the target LLM, preserving the exact output distribution. The efficiency of SD depends on the trade-off between draft generation latency and speculation accuracy, when properly balanced, SD enables the validation of multiple tokens per decoding step, making it particularly well suited for real-time and large-scale multimedia inference scenarios.

Recent developments in speculative decoding (SD) have extended the efficiency frontier by exploring diverse strategies to balance latency and accuracy. These include integrating lightweight drafting modules directly into LLM architectures~\cite{cai2024medusa,li2024eagle}, fine-tuning draft models to enhance prediction efficiency~\cite{elhoushi2024layerskip}, and aligning draft and target model distributions for improved compatibility~\cite{zhou2023distillspec}. While these approaches deliver notable acceleration gains, they often rely on additional training or specialized components, thereby limiting their generalizability and increasing deployment complexity. Another representative work SWIFT~\cite{xia2024swift}, accelerates speculative decoding through layer skipping with carefully tuned acceptance thresholds, achieving strong speedups in practice. However, despite being training-free, its deployment still involves method-task-specific optimization and manual tuning.  More broadly, the practical deployment of speculative decoding is often limited not by the decoding algorithm itself, but by the cost and complexity of acquiring, tuning, and maintaining an effective draft model. We therefore argue that draft model acquisition, rather than speculative decoding itself, constitutes the primary barrier to practical and scalable deployment.

\begin{figure*}[t]
    \vspace{-20pt}
    \centering
    \includegraphics[width=0.7\linewidth]{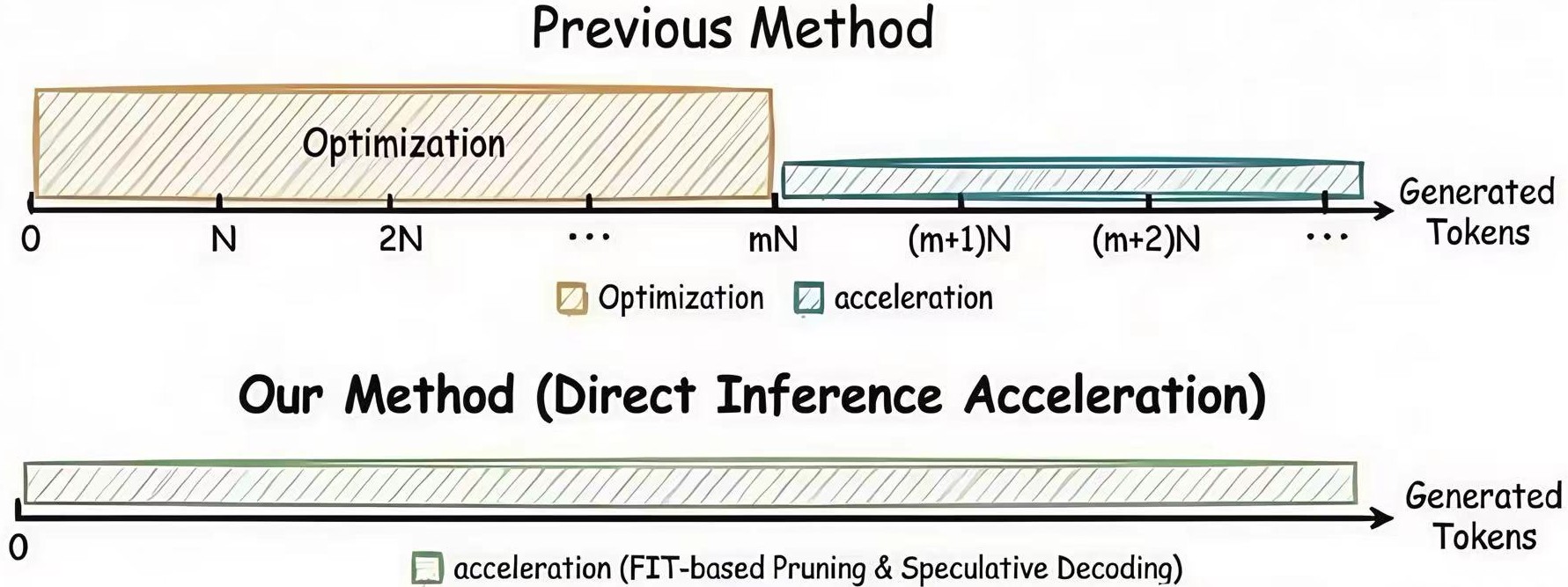}
    \caption{
    Comparison between previous optimization-based acceleration methods and our direct inference acceleration approach.
    Previous methods incur a substantial upfront optimization phase over the first $mN$ tokens before achieving limited decoding acceleration.
    In contrast, our method applies FIT-based pruning and speculative decoding directly at inference time, enabling end-to-end acceleration across the entire generation process without any offline optimization overhead.
    }
    \label{fig:direct_acceleration}
    \vspace{-10pt}
\end{figure*}

In order to address this challenge, we propose \textbf{SDFP}, a novel plug-and-play framework that integrates FIT-based pruning with speculative decoding for efficient LLM inference. SDFP leverages the Fisher Information Trace (FIT)~\cite{zandonati2022fit} to assess layer sensitivity and guide pruning \textbf{without requiring any fine-tuning or retraining}, producing a lightweight draft model for speculative decoding. Figure~\ref{fig:direct_acceleration} contrasts prior speculative decoding methods, which require iterative optimization before effective acceleration, with SDFP, which enables immediate inference acceleration by directly applying one-shot FIT-based pruning and speculative decoding, without incurring any optimization overhead. Instead of relying on heuristic or computationally expensive sensitivity estimation methods, FIT provides a principled, information-geometric measure of how parameter perturbations affect model behavior. This design enables fast, theoretically grounded, and dataset-agnostic pruning, while seamlessly supporting \textbf{inference acceleration without modifying or fine-tuning the original model}, the detailed workflow is illustrated in Figure~\ref{fig:sdfp_overview}. This design enables efficient model sparsification while preserving output fidelity, making it particularly suitable for latency-sensitive and large-scale multimedia inference scenarios. Our contributions are summarized as follows:
\begin{itemize}
    \item \textbf{ FIT-based sensitivity modeling for speculative decoding.}
    To our knowledge, we are the first to introduce \textbf{Fisher Information Trace (FIT)} as a unified sensitivity criterion for draft model construction in speculative decoding. FIT captures both parameter and activation sensitivities within a single metric, explicitly modeling how pruning-induced perturbations propagate through the network and affect the acceptance behavior of speculative decoding. Compared to conventional magnitude-based or Hessian-based criteria, FIT provides a holistic and task-agnostic measure of model robustness.

    \item \textbf{A training-free and plug-and-play pruning framework.}
   We propose \textbf{SDFP}, a fully training-free speculative decoding framework that constructs efficient draft models via FIT-based layer pruning. FIT sensitivity can be accurately estimated from a pretrained full-precision model using only a single forward–backward pass per minibatch over a small calibration set, without requiring second-order derivatives, architecture search, or task-specific fine-tuning, enabling rapid and practical deployment in interactive and multimedia systems.

    \item \textbf{Extensive evaluation across models and tasks.}
  Extensive experiments demonstrate that SDFP achieves \textbf{competitive or superior end-to-end decoding speedup} compared to existing training-dependent methods, while \textbf{incurring negligible offline overhead}. 
Across diverse models and evaluation domains, SDFP significantly improves inference efficiency and generalization, and offers a deployment-friendly, plug-and-play alternative to prior approaches.
\end{itemize}

\vspace{-2pt}
\section{Background}
\subsection{Speculative Decoding (SD)}
\label{sec:sd_background}
Speculative Decoding (SD) accelerates autoregressive generation by coupling a fast \emph{draft} model
$q_{\phi}$ with a high-quality \emph{target} model $p_{\theta}$ \cite{leviathan2023fast}. Instead of invoking
the target for every next token, SD lets the draft propose a short block of $k$ tokens and verifies
them using the target with an accept--reject rule.

Given input $x$ and an accepted prefix $y_{<t}$, the draft proposes
\begin{align}
\tilde{y}_{t:t+k-1} \sim q_{\phi}(\cdot \mid x, y_{<t}).
\label{eq:sd_draft}
\end{align}
The target then evaluates the same proposed block (efficiently via KV caching) and provides conditional
probabilities $p_{\theta}(\tilde{y}_{t+i}\mid x, y_{<t}, \tilde{y}_{t:t+i-1})$ for $i=0,\dots,k-1$.
Verification proceeds token by token with acceptance probability
\begin{align}
\alpha_i
~=~
\min\!\left(
1,\;
\frac{
p_{\theta}\!\left(\tilde{y}_{t+i}\mid x, y_{<t}, \tilde{y}_{t:t+i-1}\right)
}{
q_{\phi}\!\left(\tilde{y}_{t+i}\mid x, y_{<t}, \tilde{y}_{t:t+i-1}\right)
}
\right),
\label{eq:sd_accept}
\end{align}
where $q_{\phi}(\cdot)$ denotes the draft conditional under the same context. Accepted tokens are
committed to the output prefix. If the first rejection occurs at position $j$, SD samples the next
token from a corrected distribution:
\begin{align}
y_t \sim r_{\theta,\phi}^{(j)}(\cdot), \qquad
r_{\theta,\phi}^{(j)}(y)\propto p_{\theta}(y)-\alpha_j q_{\phi}(y),
\label{eq:sd_resample}
\end{align}
with $p_{\theta}(y)$ and $q_{\phi}(y)$ shorthand for the corresponding conditionals given
$(x, y_{<t}, \tilde{y}_{t:t+j-1})$. By construction, SD preserves the target decoding distribution
(i.e., it is lossless in distribution), while achieving speedup when the draft is accurate and many
proposed tokens are accepted.

\begin{algorithm}[t]
\caption{SDFP: Speculative Decoding with FIT-based Pruned Model}
\label{alg:sdfp}

\KwIn{
Pretrained LLM $f_\theta$ with $L$ layers; calibration dataset $\mathcal{D}$; pruning ratio $r$; speculation depth $k$; max length $L_{\max}$
}
\KwOut{Pruned draft model $\hat f$; generated sequence $y$}

\textbf{Stage A: FIT-based Layer Pruning}

Initialize FIT score $T_\ell \leftarrow 0$ for all $\ell = 1,\dots,L$\;

\ForEach{minibatch $B \subset \mathcal{D}$}{
    Compute loss $f(B,\theta)$ and gradients $\nabla_{\theta_\ell} f(B,\theta)$ for all layers\;
    \For{$\ell = 1$ \KwTo $L$}{
        $T_\ell \leftarrow T_\ell + \|\nabla_{\theta_\ell} f(B,\theta)\|_2^2$\;
    }
}
Normalize $T_\ell \leftarrow T_\ell / |\mathcal{D}|$ for all $\ell$\;

Sort layers by ascending $T_\ell$ and select prune set $\mathcal{P}$ with $|\mathcal{P}| = \lfloor rL \rfloor$\;

Construct draft model $\hat f$ by removing layers in $\mathcal{P}$\;

\vspace{0.5em}
\textbf{Stage B: Speculative Decoding with Pruned Draft Model}

Initialize output $y \leftarrow [\,]$\;

\While{not EOS and $|y| < L_{\max}$}{
    Generate $k$ speculative tokens $\hat y_{1:k}$ using draft model $\hat f$\;
    Evaluate acceptance with target model $f_\theta$\;
    Append accepted prefix of $\hat y_{1:k}$ to $y$; if none accepted, sample from $f_\theta$\;
}

\Return $\hat f, y$\;

\end{algorithm}

\section{Proposed Method}
\label{sec:method}

In this section, we introduce \textbf{SDFP} (Speculative Decoding with FIT-based Pruned Model), a training-free and plug-and-play framework that integrates \textbf{FIT-based pruning} with \textbf{speculative decoding (SD)} to achieve efficient LLM inference. 
Our method consists of two main stages: (1) pruning the model based on Fisher Information Trace (FIT) scores to construct a compact draft model, and (2) employing this pruned model within a speculative decoding paradigm for parallelized token generation and verification.

\subsection{Preliminary: Fisher Information Trace (FIT) and Sensitivity Estimation}
\label{sec:fit_background}

Quantization~\cite{su2025akvq,su2025rotatekv} and pruning introduce parameter perturbations that may degrade model performance, making layer-wise sensitivity estimation a key problem in efficient model compression.
We adopt the Fisher Information Trace (FIT)~\cite{zandonati2022fit} as a lightweight and theoretically grounded metric to characterize the sensitivity of each transformer layer to such perturbations.

\vspace{3pt}
\noindent \textbf{Information-geometric foundation.}
From the perspective of information geometry, the Fisher Information Matrix (FIM) measures how small parameter perturbations alter the model’s output distribution.
Consider a parameterized model $p(y|x, \theta)$ with perturbation $\delta\theta$.
The Kullback–Leibler divergence between the original and perturbed distributions can be approximated as:
\begin{equation}
    D_{\mathrm{KL}}(p_\theta \Vert p_{\theta+\delta\theta}) 
    = \frac{1}{2} \, \delta\theta^{\top} I(\theta) \, \delta\theta,
\end{equation}
where $I(\theta)$ denotes the Fisher Information Matrix (FIM):
\begin{equation}
    I(\theta) = \mathbb{E}_{p_\theta(x,y)} 
    \left[ \nabla_\theta \log p(y|x, \theta) 
           \nabla_\theta \log p(y|x, \theta)^{\top} \right].
\end{equation}
This establishes a natural connection between the FIM and model sensitivity: directions with high Fisher curvature correspond to parameters more critical to performance.

\begin{figure*}[t]
    \vspace{-20pt}
    \centering
    \begin{subfigure}{\textwidth}
        \centering
        \includegraphics[width=\textwidth]{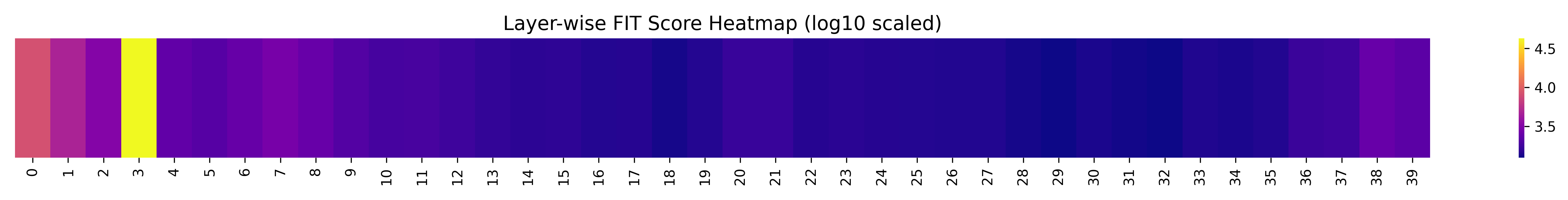}
        \label{fig:fit-13b}
        \vspace{-15pt}
        \subcaption[a]{Layer-wise FIT scores of the LLaMA-2-13B}
    \end{subfigure}
    \vspace{1em}
    \begin{subfigure}{\textwidth}
        \centering
        \includegraphics[width=\textwidth]{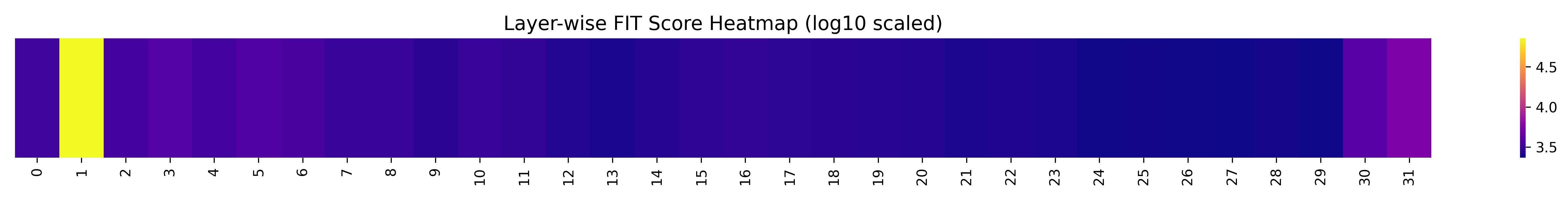}
        \label{fig:fit-7b}
        \vspace{-15pt}
        \subcaption[b]{Layer-wise FIT scores of the LLaMA-2-7B}
    \end{subfigure}
    \vspace{-5pt}
    \caption{Layer-wise FIT sensitivity heatmaps computed on WikiText2 for (a) LLaMA-2-13B and (b) LLaMA-2-7B.
    Both models exhibit non-uniform sensitivity distributions across transformer layers, indicating intrinsic layer-wise redundancy and motivating FIT-guided layer pruning in SDFP.}
    \label{fig:fit-heatmap}
    \vspace{-15pt}
\end{figure*}

\vspace{3pt}
\noindent \textbf{FIT metric and layer-wise formulation.}
Assuming quantization or pruning noise $\delta\theta$ follows a zero-mean uncorrelated distribution, the expected performance degradation can be approximated by:
\begin{equation}
    \Omega = \operatorname{Tr}\!\left(I(\theta) \, 
    \mathrm{diag}\!\left(\mathbb{E}[\delta\theta^2]\right)\right).
\end{equation}
Aggregating this effect across network layers yields the layer-wise FIT form:
\begin{equation}
    \Omega = \sum_{l=1}^{L} 
    \operatorname{Tr}\!\left(I(\theta_l)\right) 
    \, \mathbb{E}[\delta\theta_l^2],
\end{equation}
where $\operatorname{Tr}(\cdot)$ denotes the matrix trace and $\mathbb{E}[\delta\theta_l^2]$ models the perturbation magnitude induced by quantization or pruning in layer $l$.

\vspace{3pt}
\noindent \textbf{Computational efficiency.}
Direct computation of the Hessian trace is often prohibitive for large-scale models. 
FIT instead leverages the \textit{Empirical Fisher} approximation:
\begin{equation}
    \widehat{I}(\theta) = 
    \frac{1}{N}\sum_{i=1}^{N} 
    \nabla_\theta f(z_i, \theta) 
    \nabla_\theta f(z_i, \theta)^{\top},
\end{equation}
where $f(z_i, \theta)$ denotes the per-sample loss. 
Its trace can be computed efficiently as:
\begin{equation}
    \operatorname{Tr}\!\left[\widehat{I}(\theta)\right]
    = \frac{1}{N}\sum_{i=1}^{N} 
    \|\nabla_\theta f(z_i, \theta)\|_2^2.
\end{equation}
This formulation requires only a single forward–backward pass per minibatch over a small calibration set, eliminating the need for second-order derivatives while providing low-variance, model-agnostic estimates. 
Empirical results demonstrate that the variance of this estimator is orders of magnitude smaller than that of Hutchinson-based Hessian trace estimators, resulting in substantially faster convergence.

\vspace{3pt}
\noindent \textbf{Joint sensitivity of weights and activations.}
Unlike previous metrics that treat weight and activation perturbations separately, FIT unifies both contributions under the same geometric framework.
By extending the parameter manifold to include activation statistics $\hat{a}$, the perturbed model is written as $p(y|x, \theta+\delta\theta, \hat{a}+\delta\hat{a})$, 
and the combined layer-wise FIT can be expressed as:
\begin{equation}
    \Omega_{\mathrm{FIT}} = 
    \sum_{l=1}^{L}
    \left(
    \operatorname{Tr}\!\left(I(\theta_l)\right)\,
    \mathbb{E}[\delta\theta_l^2]
    +
    \operatorname{Tr}\!\left(I(\hat{a}_l)\right)\,
    \mathbb{E}[\delta\hat{a}_l^2]
    \right).
\end{equation}
This joint treatment captures how both parameter and activation quantization (or pruning) influence the network’s robustness, 
yielding a single, unified sensitivity score that correlates strongly with post-compression accuracy.
In pruning, we assume the perturbation variance is approximately layer-wise constant.
With this assumption, FIT is proportional to the \emph{empirical Fisher trace}, which we adopt as the practical layer sensitivity score in Eq.~(9).

Overall, FIT provides a \textit{fast}, \textit{training-free}, and \textit{theoretically
justified} criterion for layer importance estimation, forming the
foundation for our FIT-based pruning strategy.

\subsection{FIT-based Pruning}
To begin, we utilize the general-purpose \textit{WikiText2} dataset to compute the FIT scores for each transformer layer. 
The FIT score provides a theoretically grounded measure of layer sensitivity by quantifying how parameter perturbations affect the model output distribution (see Section~\ref{sec:fit_background}). 
For each layer $l$, the empirical Fisher trace is computed as:
\begin{equation}
    T_l = \operatorname{Tr}\!\left[\widehat{I}(\theta_l)\right] 
    = \frac{1}{N} \sum_{i=1}^{N} 
    \|\nabla_{\theta_l} f(z_i, \theta)\|_2^2,
\end{equation}
where $f(z_i, \theta)$ denotes the loss function evaluated on sample $z_i = (x_i, y_i)$, and $\widehat{I}(\theta_l)$ is the empirical Fisher information of layer $l$. 
Larger FIT values indicate higher sensitivity, implying that pruning such layers would result in greater performance degradation.

We rank all layers based on their FIT scores and prune those with the lowest sensitivity, guided by pruning ratios derived from prior Bayesian learning-based pruning methods.
This strategy balances compression and model fidelity while ensuring theoretical soundness.
The resulting pruned network serves as a lightweight \textit{draft model} that preserves the most informative layers of the original architecture. Figure~\ref{fig:fit-heatmap} visualizes the layer-wise FIT sensitivity of LLaMA-2-13B. Due to the large dynamic range of raw FIT values across layers, we apply a logarithmic transformation when visualizing FIT scores. This log-scale representation improves interpretability by preventing a small number of highly sensitive layers from dominating the visualization.
We observe substantial variation across transformer layers, indicating pronounced redundancy in specific regions of the network.
This observation motivates our FIT-guided layer pruning strategy in SDFP.


\begin{table*}[t]
\vspace{-10pt}
\centering
\small
\setlength{\tabcolsep}{4.5pt}
\renewcommand{\arraystretch}{1.2}
\begin{tabular}{l l c c c c c c c c}
\toprule
\textbf{Models} & \textbf{Methods} &
\multicolumn{2}{c}{\textbf{CNN/DM}} &
\multicolumn{2}{c}{\textbf{GSM8K}} &
\multicolumn{2}{c}{\textbf{TinyStories}} &
\textbf{Speed} & \textbf{Overall} \\
& & $\alpha$ & Speedup & $\alpha$ & Speedup & $\alpha$ & Speedup &
(tokens/s) & Speedup \\
\midrule
\multirow{5}{*}{LLaMA-2-13B} 
& VANILLA & N/A & $1.00\times$ & N/A & $1.00\times$ & N/A
& $1.00\times$ & 20.10 & $1.00\times$ \\
& PARALLEL & N/A & $0.95\times$ & N/A & $0.99\times$ & N/A & $0.97\times$ & 19.49 & $0.97\times$ \\
& LOOKAHEAD & N/A & $1.16\times$ & N/A & $1.29\times$ & N/A & $1.35\times$ & 25.46 & $1.27\times$ \\
& SWIFT & 0.99 & $1.37\times$ & 0.98 & $1.31\times$ & 1.00 & $1.53\times$ & 28.26 & $1.41\times$ \\
& Ours & 0.95 & $1.32\times$ & 0.95 & $1.36\times$ & 0.98 & $1.49\times$ & 28.05 & $1.40\times$ \\
\midrule

\multirow{5}{*}{LLaMA-2-13B-Chat}
& VANILLA & N/A & $1.00\times$ & N/A & $1.00\times$ & N/A & $1.00\times$ & 19.96 & $1.00\times$ \\
& PARALLEL & N/A & $0.96\times$ & N/A & $0.97\times$ & N/A & $0.98\times$ & 19.26 & $0.97\times$ \\
& LOOKAHEAD & N/A & $1.15\times$ & N/A & $1.31\times$ & N/A & $1.40\times$ & 25.69 & $1.29\times$ \\
& SWIFT & 0.90 & $1.28\times$ & 0.92 & $1.25\times$ & 0.99 & $1.50\times$ & 26.80 & $1.34\times$ \\
& Ours & 0.92 & $1.23\times$ & 0.92 & $1.27\times$ & 0.95 & $1.44\times$ & 26.21 & $1.31\times$ \\
\midrule

\multirow{5}{*}{LLaMA-2-70B}
& VANILLA & N/A & $1.00\times$ & N/A & $1.00\times$ & N/A & $1.00\times$ & 4.32 & $1.00\times$ \\
& PARALLEL & N/A & $0.95\times$ & N/A & $0.97\times$ & N/A & $0.96\times$ & 4.14 & $0.96\times$ \\
& LOOKAHEAD & N/A & $1.15\times$ & N/A & $1.35\times$ & N/A & $1.35\times$ & 5.45 & $1.26\times$ \\
& SWIFT & 0.99 & $1.43\times$ & 0.98 & $1.39\times$ & 0.99 & $1.62\times$ & 6.41 & $1.48\times$ \\
& Ours & 0.96 & $1.45\times$ & 0.95 & $1.37\times$ & 0.95 & $1.55\times$ & 6.29 & $1.46\times$ \\
\bottomrule
\end{tabular}
\caption{
Comparison between SDFP and prior plug-and-play methods.  We report the acceptance rate $\alpha$, speedup ratio, and average decoding speed (tokens/s) under greedy decoding. \textbf{Notably, SDFP requires no auxiliary training, Bayesian optimization, or hyperparameter tuning, while achieving performance comparable to prior state-of-the-art methods.}
}
\label{tab:main_results}
\vspace{-15pt}
\end{table*}

\noindent \textbf{Key Advantage.} 
Unlike existing pruning approaches that require task-specific datasets or Bayesian optimization for layer selection, SDFP computes FIT scores using a general dataset (\textit{WikiText2}) that generalizes across tasks. 
This eliminates the need for retraining or task-dependent adjustment of pruning configurations, making our method fully \textit{plug-and-play} and easy to deploy in diverse scenarios.

\subsection{Speculative Decoding with FIT-based Draft Model}
After pruning, the compact model is employed as the \textit{draft model} in the speculative decoding framework, while the original full LLaMA-2 model acts as the \textit{target model}. 
At each decoding step, the draft model predicts multiple speculative tokens in parallel, which are then verified by the target model to ensure consistency with its output distribution. 
This two-stage process effectively decouples generation and verification, allowing SDFP to achieve faster decoding while maintaining identical generation quality.

Formally, let $q_{\phi}$ and $p_{\theta}$ denote the draft and target token distributions, respectively.
At each step, the draft proposes a $k$-token block and the target verifies it using the standard accept--reject rule
(Section~II-A). Specifically, each proposed token is accepted with probability
$\alpha_i=\min\!\left(1,\frac{p_{\theta}}{q_{\phi}}\right)$ under the same context, and decoding commits the longest accepted prefix.
If a rejection occurs, the next token is sampled from the corrected distribution as in Eq.~(3), after which the process continues.
Since our draft model is obtained by structured layer pruning from the same pretrained LLM, it remains architecture-compatible with
the target, enabling efficient KV reuse during verification and typically yielding high acceptance rates in practice.

\subsection{Overall Framework}
The overall workflow of SDFP is illustrated in Algorithm~1. 
Specifically, SDFP first computes layer-wise FIT scores on a general corpus to quantify sensitivity, and then prunes the least informative transformer layers according to a target pruning ratio. 
The resulting pruned model is used as a lightweight \textit{draft model}, while the original full model serves as the \textit{target model}. 
During inference, speculative decoding is performed by generating speculative tokens with the draft model and verifying them in parallel using the target model.

This unified framework tightly integrates model compression and decoding acceleration into a single, training-free procedure. 
By leveraging intrinsic layer redundancy revealed by FIT-based sensitivity analysis, SDFP enables efficient LLM inference with no retraining, no auxiliary modules, and full compatibility with existing speculative decoding pipelines.

\section{Experiments}
\subsection{Experimental Setup}

\begin{table*}[t]
\vspace{-7pt}
\centering
\begin{tabular}{lcc}
\toprule
\textbf{Methods} & \textbf{Training Cost} & \textbf{Optimization Latency} \\
\midrule
LayerSkip-SD & $50 \times 10^{3}$ training steps with 64 A100 (80GB) & N/A \\
SELF-SD & 1000 Bayesian Optimization Iterations Before inference & $\sim$7.5 hours \\
SWIFT & N/A & $\sim$A few minutes each time \\
\textbf{SDFP (Ours)} & \textbf{N/A} & \textbf{N/A} \\
\bottomrule
\end{tabular}
\caption{Comparison of speculative decoding methods in terms of draft model construction overhead for LLaMA-2-13B.
Training cost is reported from the original papers, while optimization latency is measured on a single A6000 GPU.}
\label{tab:draft_overhead}
\vspace{-10pt}
\end{table*}

\paragraph{Implementation Details}
We evaluate our proposed method, \textbf{SDFP}, on the \textbf{LLaMA-2} family of models across a variety of representative tasks, including summarization, mathematical reasoning, and storytelling. The evaluation covers three benchmark datasets: \textbf{CNN/Daily Mail (CNN/DM)}~\cite{nallapati2016abstractive} for summarization, \textbf{GSM8K}~\cite{cobbe2021training} for reasoning, and \textbf{TinyStories}~\cite{eldan2023tinystories} for narrative generation. This diverse task suite enables a comprehensive evaluation of both the efficiency and generality of our approach.
For generation settings, the maximum output lengths are set to 64 for CNN/DM and GSM8K, and 128 for TinyStories. We adopt a 1-shot evaluation protocol for CNN/DM and TinyStories, and a 5-shot setting for GSM8K, following standard practice in prior works~\cite{leviathan2023fast,xia2023speculative}. For each dataset, we randomly sample 1,000 test instances from the evaluation set. All decoding experiments are performed with a batch size of 1 to ensure consistency in latency measurement.
During pruning, FIT scores are computed using the general-purpose WikiText2 dataset, and layers are ranked based on their sensitivity. The pruning ratio for each model follows the layer-sparsity proportions derived from Bayesian learning–based pruning methods. 
For example, in LLaMA-2-13B, we prune 50\% of attention layers and 35\% of FFN layers, a configuration identified through extensive empirical evaluation as providing the best accuracy–efficiency trade-off. The resulting pruned model is employed as the \textit{draft model} in the speculative decoding process, while the original LLaMA-2 model serves as the \textit{target model}. All experiments are conducted under the speculative decoding framework with identical generation parameters to ensure fair comparison. 
\paragraph{Baselines}
In our main experiments, we compare \textbf{SDFP} against several representative \textit{training-free speculative decoding} methods. Specifically, we include two Jacobi-based plug-and-play approaches—\textbf{Parallel Decoding}~\cite{santilli2023accelerating} and \textbf{Lookahead Decoding}~\cite{fu2024break}—which leverage iterative pseudo-token refinement for efficient drafting. In addition, we compare against \textbf{SWIFT}~\cite{xia2024swift}, a recent layer-skipping SD framework that dynamically selects subsets of layers during inference. 
It is worth noting that SDFP and SWIFT are conceptually related yet orthogonal: while SWIFT explores dynamic layer selection through Bayesian optimization, SDFP employs FIT-based layer sensitivity to construct a pruned draft model in a fully plug-and-play manner without any task-specific retraining or optimization. For fairness, all methods are evaluated under identical decoding configurations and model architectures. Other speculative decoding methods that require additional auxiliary networks or fine-tuning (e.g., EAGLE, Medusa) are excluded from comparison, as they rely on extra training stages that limit general applicability.\par\allowbreak
\paragraph{Evaluation Metrics}
We evaluate \textbf{SDFP} using several widely adopted metrics to measure inference efficiency and decoding performance. Specifically, we report the \textit{token acceptance rate} $\alpha$, which together characterize the behavior of speculative decoding. In addition to these standard SD metrics, we also measure the actual \textit{decoding throughput} (tokens per second) and the \textit{wall-time speedup ratio} relative to vanilla autoregressive decoding. Since speculative decoding theoretically guarantees that the output distribution of the target LLM remains unchanged, evaluating generation quality is generally unnecessary. 

\subsection{Main Results}

Table~\ref{tab:draft_overhead} compares SDFP with prior speculative decoding methods in terms of draft model construction overhead. Although several existing approaches are training-free, they still incur non-negligible optimization latency due to threshold tuning or Bayesian search procedures before inference. In contrast, SDFP introduces no optimization phase and constructs the draft model through a one-shot FIT-based pruning step, which makes SDFP a deployment-friendly method. As a result, SDFP achieves the lowest end-to-end preparation cost, enabling rapid local deployment without additional training or optimization.

Table~\ref{tab:main_results} presents the comparison between SDFP and existing plug-and-play speculative decoding methods across all text generation benchmarks. 
The experimental results reveal the following observations: 

(1) SDFP achieves comparable acceleration performance to previous state-of-the-art methods, maintaining similar decoding speedups over vanilla autoregressive decoding across different tasks and model scales. 
Despite \textbf{not relying on any task-specific optimization or retraining}, SDFP attains consistent improvements in decoding throughput, demonstrating the effectiveness of FIT-based pruning for speculative decoding. 

(2) Unlike prior methods such as SWIFT or Jacobi-based speculative decoding, which require Bayesian optimization or task-specific layer configuration, SDFP is fully \textbf{training-free and plug-and-play}. 
By computing FIT scores on a general-purpose dataset (WikiText2), our approach can be directly applied to new tasks or domains without additional tuning or optimization steps. 
This substantially simplifies deployment while preserving inference efficiency comparable to SOTA approaches.

(3) The stable acceleration and high acceptance rates indicate that the pruned draft retains sufficient representational capacity for effective speculation.Moreover, SD verification preserves the target decoding distribution by design, ensuring faithful decoding behavior.
Overall, SDFP offers stable, plug-and-play inference acceleration and consistently achieves strong speedups across benchmarks.

\vspace{-3pt}
\section{Conclusion}

We present \textbf{SDFP (Speculative Decoding with FIT-based Pruned Models)}, a fully training-free and plug-and-play framework for accelerating large language model (LLM) inference via speculative decoding. SDFP addresses the practical bottleneck of draft model acquisition by constructing an efficient draft directly from a pretrained LLM using \textbf{Fisher Information Trace (FIT)} layer sensitivity estimation, requiring no retraining, task-specific tuning, or architecture search. By pruning low-impact layers while preserving architectural compatibility, SDFP integrates seamlessly into standard speculative decoding. Experiments across multiple models and benchmarks show \textbf{1.32$\times$--1.5$\times$ speedup}, while SD verification \textbf{preserves the target decoding distribution by design}.






\bibliographystyle{IEEEtran}
\bibliography{icme2026references}


\end{document}